\documentclass[runningheads]{llncs}
\usepackage[T1]{fontenc}
\usepackage{graphicx,verbatim}
\usepackage{amsmath}
\usepackage{graphicx}%
\usepackage{multirow}%
\usepackage{amsmath,amssymb,amsfonts}%

\usepackage{mathrsfs}%
\usepackage[title]{appendix}%
\usepackage{xcolor}%
\usepackage{textcomp}%
\usepackage{manyfoot}%
\usepackage{booktabs}%
\usepackage{algorithm}%
\usepackage{algorithmicx}%
\usepackage{algpseudocode}%
\usepackage{listings}%
\usepackage{cleveref}
\usepackage{subcaption}
\begin{document}
\title{All-in-One Augmented Reality Guided Head and Neck Tumor Resection}
%
 
\author{
Yue Yang\inst{1} \and
Matthieu Chabanas\inst{1} \and
Carrie Reale\inst{2} \and
Annie Benson\inst{1} \and
Jason Slagle\inst{2} \and
Matthew Weinger\inst{2} \and
Michael Topf\inst{2} \and
Jie Ying Wu\inst{1}
}

\institute{
Vanderbilt Institute for Surgery and Engineering, Nashville, TN 37235, USA
\and
Vanderbilt University Medical Center, Nashville, TN 37232, USA
}

\maketitle              

\begin{abstract}
Positive margins are common in head and neck squamous cell carcinoma, yet intraoperative re-resection is often imprecise because margin locations are typically communicated verbally from pathology. We present an all-in-one augmented reality (AR) system that relocalizes positive margins from a resected specimen to the resection bed and visualizes them in situ using HoloLens~2 depth sensing and fully automated markerless surface registration. In a silicone phantom study with six medical trainees, markerless registration achieved target registration errors comparable to a marker-based baseline (median $1.8$~mm vs.\ $1.7$~mm; maximum $<4$~mm). In a margin relocalization task, AR guidance reduced error from verbal guidance (median $14.2$~mm) to a few millimeters (median $3.2$~mm), with all AR localizations within 5~mm error. These results support the feasibility of markerless AR margin guidance for more precise intraoperative re-excision.
\end{abstract}

\section{Introduction}
Head and neck squamous cell carcinoma (HNSCC) affects hundreds of thousands of patients each year \cite{barsouk2023epidemiology} and has one of the highest rates of positive surgical margins among solid tumors \cite{orosco2018positive}, contributing to local recurrence and poor outcomes. Intraoperative frozen section analysis is used in over 90\% of cases \cite{long2022use}, and a positive margin requires the surgeon to re-resect accurately by relocating its position. This is challenging due to complex anatomy and specimen disorientation. In practice, pathologists convey margin locations verbally, which provides only coarse spatial guidance \cite{nakhleh2011quality}; consequently, re-resections are often imprecise, and only about 20--30\% of re-resection specimens contain residual tumor \cite{prasad2024often}.

Augmented reality (AR) head-mounted displays can overlay margin information directly in the surgical field \cite{tong2024development}. Prior pilots demonstrated feasibility but relied on manual alignment \cite{necker2023virtual,tong2024development} or marker-based registration \cite{duan2025localization}, which can add workflow burden and hinder deployment. Markerless registration instead aligns to anatomy using depth sensing and computer vision; it has achieved few-millimeter accuracy for rigid structures \cite{ury2025markerless,nasir2023augmented}, though head and neck surfaces remain challenging.

Here, we present a markerless, automated AR margin guidance system that registers the resected specimen back to the resection bed using only geometric data and visualizes margin locations in situ. We evaluate its performance on a silicone face phantom with quantitative registration error measurements through a pilot user study (five residents, one medical student), comparing AR-guided versus verbal margin relocation.

\section{Methods}
\begin{figure}[t]
    \centering
    \includegraphics[width=1\linewidth]{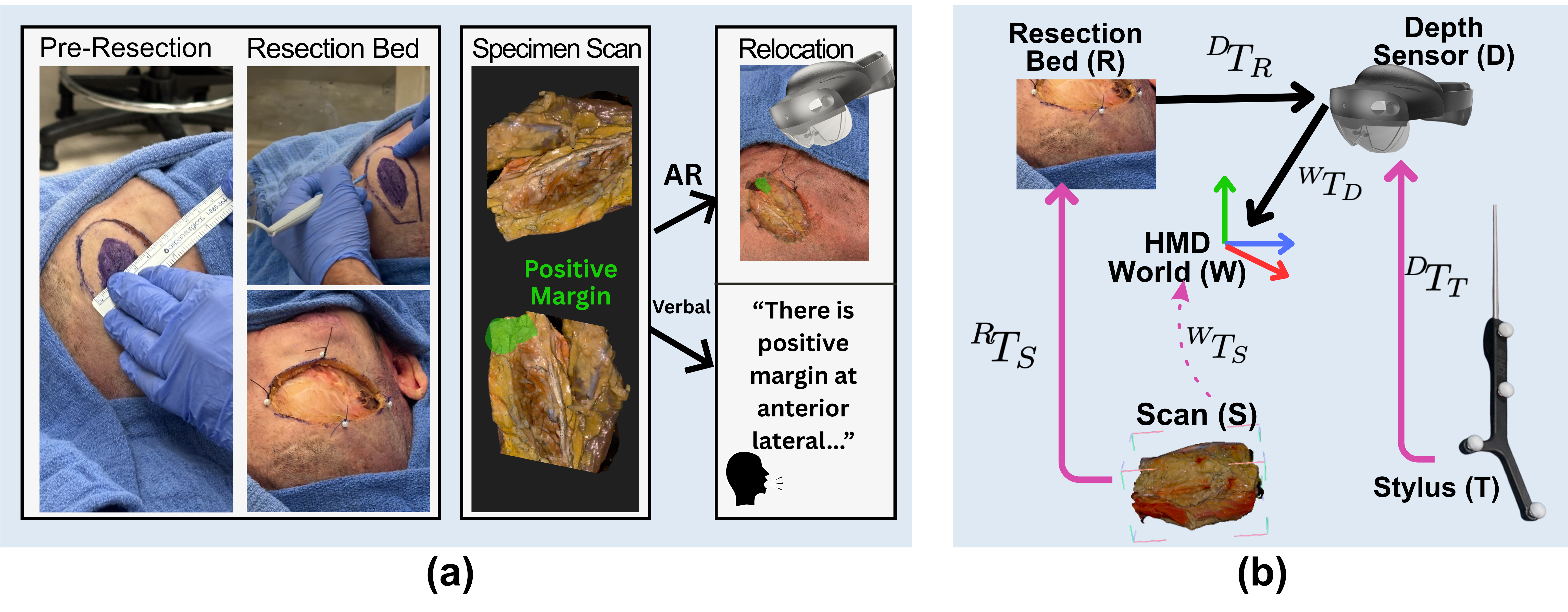}
    \caption{System pipeline and coordinates: (a) AR-guided vs.\ traditional margin relocalization workflow. (b) System formulation.}
    \label{fig:pipeline}
\end{figure}

As shown in Figure~\ref{fig:pipeline}(a), in a phantom scenario, our system relocalizes a positive margin by scanning the specimen, capturing the resection bed with the HoloLens~2 Articulated HAndTracking (AHAT) depth sensor, performing fully automated markerless registration, and rendering the annotated margin in AR. The AHAT sensor also continuously tracks a stylus for quantitative evaluation.

We define right-handed coordinate frames: Head-Mounted Display (HMD) world $W$, AHAT depth sensor $D$, resection bed $R$, specimen scan $S$, and tool $T$. A rigid transform from frame $X$ to $Y$ is ${}^{Y}\!T_X\in SE(3)$.

The AR overlay requires placing the specimen scan in the world frame:
\begin{equation}
{}^{W}\!T_{S} = {}^{W}\!T_{D}\;{}^{D}\!T_{R}\;{}^{R}\!T_{S}
\quad \Leftrightarrow \quad
{}^{D}\!T_{S} = \underbrace{{}^{D}\!T_{R}}_{T_{DR}}\;\underbrace{{}^{R}\!T_{S}}_{T_{RS}},
\label{eq:chain}
\end{equation}
${}^{W}\!T_D$ is provided by HoloLens tracking; ${}^{D}\!T_R$ is known from our depth-fusion reference; and the only unknown is ${}^{R}\!T_S$, estimated by markerless surface registration. Any annotated margin point $\mathbf{x}_S$ is visualized by $\mathbf{x}_W = {}^{W}\!T_S\,\tilde{\mathbf{x}}_S$.

\subsection{Resection-bed and specimen surface acquisition}
\textbf{Resection bed.} The surgeon or resident looks at the exposed cavity for $\sim$3\,s while AHAT depth frames are streamed. Each frame is back-projected to 3D and fused into a fixed resection-bed reference frame $R$ using the HMD pose, followed by voxel downsampling and outlier removal to obtain a denoised bed cloud
$\mathcal{P}_R=\{\mathbf{p}_i\}_{i=1}^{N_R}$ with normals $\mathbf{n}_{\mathbf{p}_i}$.

\textbf{Specimen scan.} The excised specimen is 3D scanned to obtain a point cloud (or mesh sampled to points)
$\mathcal{P}_S=\{\mathbf{q}_j\}_{j=1}^{N_S}$ in frame $S$, along with a margin annotation defined on the scan.

\subsection{Fully automated markerless specimen-to-bed registration}
We eliminate the manual ROI selection required by prior work \cite{yang2025markerless} by (i) improving global registration robustness for small targets (2--4\,cm specimens) and (ii) automatically constructing a tight refinement ROI from the coarse pose.

\subsubsection{Coarse global registration}
We compute local curvature scores $k(\cdot)$ on $\mathcal{P}_R$ and $\mathcal{P}_S$, and bias keypoint sampling toward landmark-rich regions:
\begin{equation}
\mathbb{P}(\mathbf{p}_i)\propto \frac{k(\mathbf{p}_i)}{\sum_{m=1}^{N_R}k(\mathbf{p}_m)}.
\label{eq:curv}
\end{equation}
We compute FPFH descriptors $\phi(\cdot)$ at sampled points and generate putative correspondences by nearest-neighbor matching in descriptor space \cite{rusu2009fast}:
\begin{equation}
\mathcal{C}=\{(i,j)\;|\;\|\phi(\mathbf{p}_i)-\phi(\mathbf{q}_j)\|<\tau\}.
\label{eq:corr}
\end{equation}
To tolerate the high outlier rates typical for small anatomical surfaces, we estimate ${}^{R}\!T_S$ using a TEASER++-style truncated least-squares objective with correspondence weights $w_{ij}$ \cite{yang2020teaser}:
\begin{equation}
\min_{\mathbf{R}\in SO(3),\,\mathbf{t}\in\mathbb{R}^3}
\sum_{(i,j)\in\mathcal{C}} w_{ij}\;
\min\!\left(\big\|\mathbf{p}_i-(\mathbf{R}\mathbf{q}_j+\mathbf{t})\big\|^2,\;\epsilon^2\right).
\label{eq:tls}
\end{equation}
As in TEASER++, we use translation-invariant pairwise differences to decouple translation from rotation:
\begin{equation}
\Delta\mathbf{p}_{ik}=\mathbf{p}_i-\mathbf{p}_k,\qquad
\Delta\mathbf{q}_{j\ell}=\mathbf{q}_j-\mathbf{q}_\ell,
\qquad
\Delta\mathbf{p}_{ik}\approx \mathbf{R}\Delta\mathbf{q}_{j\ell}\;\;\text{(inliers)}.
\label{eq:ti}
\end{equation}
This yields a coarse estimate ${}^{R}\!T_{S}^{\mathrm{coarse}}$ without any user input.

\subsubsection{Automatic ROI construction}
Instead of a user-defined ROI, we gate refinement using the coarse pose and improved prior work by \cite{yang2025markerless}. Let $\hat{\mathbf{q}}_j=\mathbf{R}^{\mathrm{coarse}}\mathbf{q}_j+\mathbf{t}^{\mathrm{coarse}}$ be specimen points in $R$. We form a dilated bounding box
\begin{equation}
\mathcal{B}=\mathrm{Dilate}\!\big(\mathrm{AABB}(\{\hat{\mathbf{q}}_j\}),\,m\big),
\end{equation}
and automatically crop the bed points for ICP:
\begin{equation}
\mathcal{P}_R^{\mathrm{ICP}}=\{\mathbf{p}\in\mathcal{P}_R\;|\;\mathbf{p}\in\mathcal{B}\}.
\label{eq:autoROI}
\end{equation}
This focuses computation on the small target structure and prevents false matches elsewhere in the scene, without any manual selection.

\subsubsection{Fine registration}
Starting from ${}^{R}\!T_{S}^{\mathrm{coarse}}$, we refine with point-to-plane ICP on $\mathcal{P}_R^{\mathrm{ICP}}$ \cite{zhang2021fast}:
\begin{equation}
\min_{\mathbf{R}\in SO(3),\,\mathbf{t}\in\mathbb{R}^3}
\sum_{(\mathbf{p},\mathbf{q})\in\mathcal{C}_k}
w_{\mathbf{p}\mathbf{q}}
\Big[\big(\mathbf{R}\mathbf{q}+\mathbf{t}-\mathbf{p}\big)^\top \mathbf{n}_{\mathbf{p}}\Big]^2,
\label{eq:icp}
\end{equation}
with robust weights and outlier rejection. The output is ${}^{R}\!T_S^{\mathrm{final}}\equiv T_{RS}$, used in \eqref{eq:chain} for AR overlay of the specimen and its annotated positive margin.

We further track a stylus/tool instrumented with retro-reflective markers using the STTAR algorithm \cite{martin2023sttar}. STTAR estimates the tool pose in the AHAT sensor frame as
$T_{TS}\equiv{}^{D}\!T_T$ at runtime. Continuous tool tracking in the world frame is obtained by composing with HMD tracking:
\begin{equation}
T_{TW}\equiv{}^{W}\!T_T = {}^{W}\!T_D\;{}^{D}\!T_T.
\label{eq:tool}
\end{equation}
STTAR reports sub-millimeter accuracy, enabling reliable quantitative evaluation of margin relocalization.
\begin{figure}[t]
    \centering
    \includegraphics[width=1\linewidth]{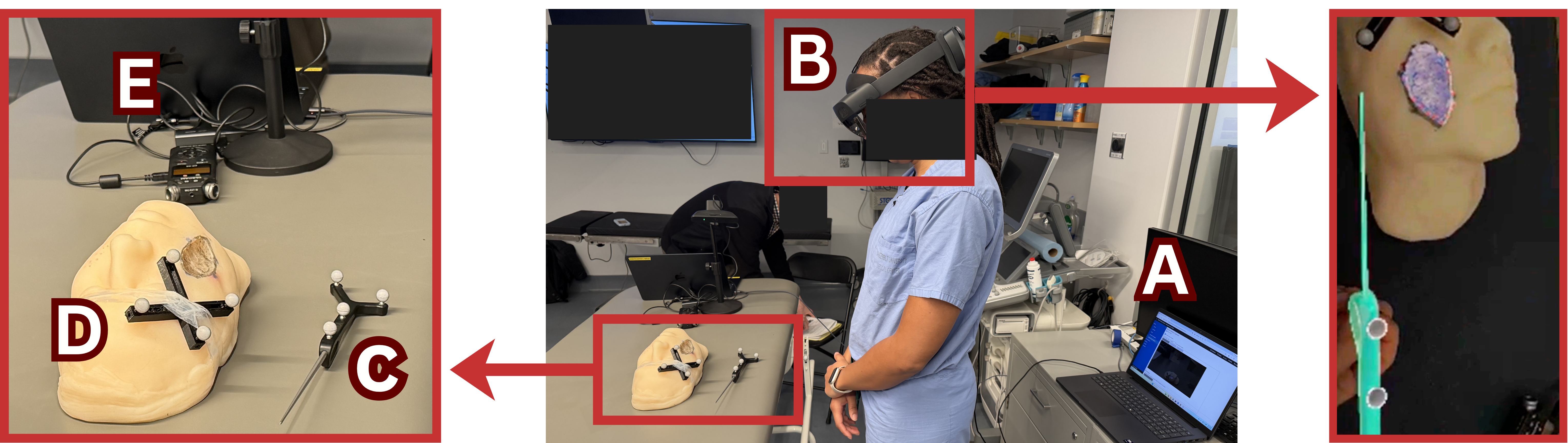}
    \caption{User study setup: (A) Mixed-reality capture, (B) HMD, (C) tracked stylus, (D) phantom (marker + markerless), (E) close-up mic/camera.}
    \label{fig:setup}
\end{figure}

\section{Results}
As shown in Figure~\ref{fig:setup}, we evaluated the system on a silicone head-and-neck phantom using a pilot study with six participants. The phantom contained an artificial cheek tumor that was resected to create a wound cavity and specimen. A structured-light scanner was used to acquire a 3D model of the specimen for AR visualization. For baseline comparison, an optical marker was attached to enable conventional marker-based registration.

\begin{figure}[t]
    \centering
    \includegraphics[width=1\linewidth]{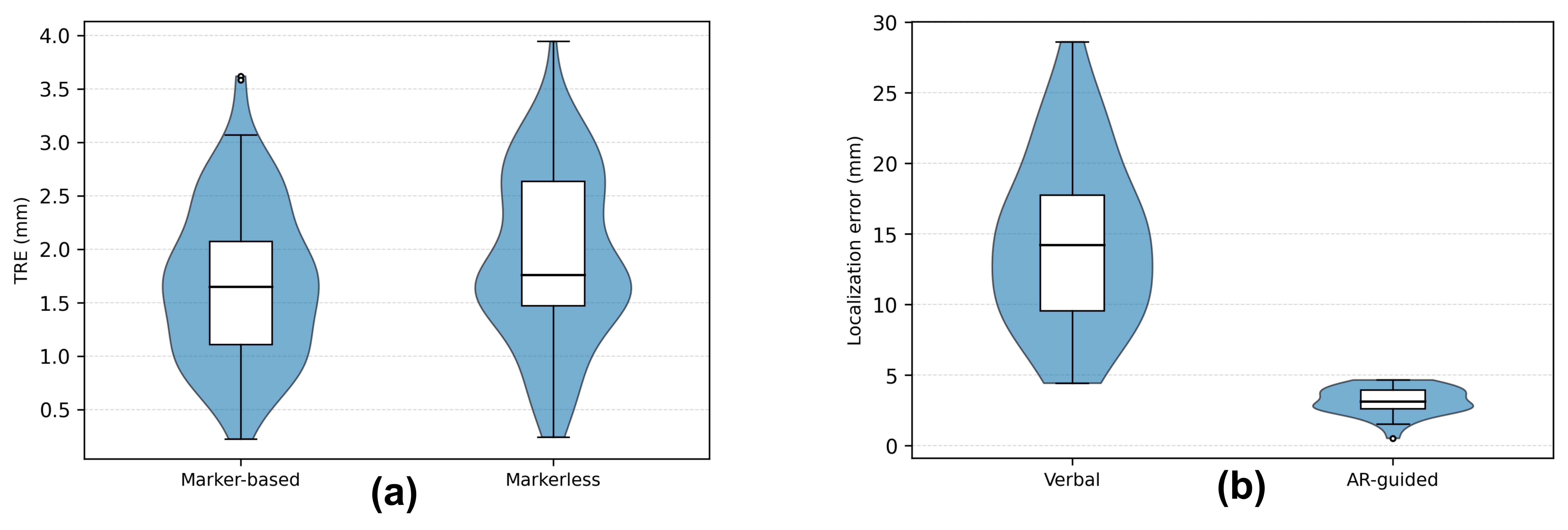}
    \caption{(a) TRE for marker vs.\ markerless registration. (b) Margin relocalization error for verbal vs.\ AR guidance.}
    \label{fig:fig1_violin}
\end{figure}

Participants wore a Microsoft HoloLens~2 and completed two tasks under two registration conditions (marker-based vs.\ markerless), with condition order counterbalanced. In the \textbf{point matching task}, 19 red holographic targets were displayed on the resection surface; participants touched each target with a tracked stylus, and the system logged the 3D contact points. Target registration error (TRE) was computed as the distance between the indicated and true target locations, with ground truth obtained by an expert touching each target twice and averaging. Each participant performed the task once per registration condition.

In the \textbf{margin relocation task}, participants localized six simulated positive margins using (1) verbal guidance mimicking a pathology report and (2) AR guidance showing a 5~mm green target at the margin location. Participants aligned the stylus tip to the estimated (verbal) or displayed (AR) margin and said “aligned,” triggering 3D logging; expert measurements provided ground truth. The task was repeated under both registration conditions, yielding 12 verbal and 12 AR localizations per participant.
\subsubsection{Registration accuracy}
Participants were five ENT trainees (PGY-1 to PGY-5 plus) and one medical student, mean age 30.3 years (range 27--33) with 4/6 female. All five trainees reported low prior VR/AR experience, and the medical student had used AR in work scenario. Using the point matching task, we computed target registration error (TRE) for each of the $19$ landmarks per user ($6$ users), yielding $114$ point targets per registration method. Figure~\ref{fig:fig1_violin}(a) shows the TRE distributions under marker-based and markerless registration. Across all points, marker-based registration achieved a TRE median of $1.7$~mm (Q1--Q3: $1.1$--$2.1$~mm), while markerless registration achieved a TRE median of $1.8$~mm (Q1--Q3: $1.5$--$2.6$~mm) (Table~\ref{tab:big_results_table}). Both methods produced maximum errors under $4$~mm (Hausdorff: $3.6$~mm marker-based, $3.9$~mm markerless). Treating each participant as an independent sample ($n=6$) and comparing participant-level median TRE, markerless registration did not differ significantly from marker-based registration (paired Wilcoxon signed-rank test, $p=0.31$). These results indicate that the automated depth-only registration achieves accuracy comparable to a fiducial baseline for this facial specimen setup.

\begin{table}[t]
\centering
\scriptsize
\setlength{\tabcolsep}{2.0pt}
\caption{\textbf{Participant-level and group-level performance.}
Distances are in mm and Cov$_5$ is in \%.}
\label{tab:big_results_table}

\begin{subtable}{\linewidth}
\centering
\caption{Registration (point matching): marker-based vs. markerless.}
\resizebox{\linewidth}{!}{%
\begin{tabular}{l|cccc|cccc}
\hline
 & \multicolumn{4}{c|}{Marker-based reg.} &
   \multicolumn{4}{c}{Markerless reg.} \\
User &
Med [Q1,Q3] & RMSE & Haus & Cov$_5$ &
Med [Q1,Q3] & RMSE & Haus & Cov$_5$ \\
\hline
P1 & 2.4 [1.8,2.7] & 2.4 & 3.6 & 100.0 & 2.5 [2.1,2.8] & 2.5 & 3.5 & 100.0 \\
P2 & 1.3 [1.1,1.6] & 1.4 & 2.1 & 100.0 & 1.7 [1.6,2.1] & 1.9 & 2.8 & 100.0 \\
P3 & 1.0 [0.8,1.4] & 1.1 & 1.7 & 100.0 & 1.4 [1.1,1.6] & 1.4 & 2.1 & 100.0 \\
P4 & 1.9 [1.4,2.0] & 1.8 & 2.6 & 100.0 & 2.9 [2.5,3.1] & 2.9 & 3.9 & 100.0 \\
P5 & 1.1 [0.7,1.6] & 1.2 & 2.1 & 100.0 & 1.6 [1.2,1.7] & 1.6 & 2.7 & 100.0 \\
P6 & 2.4 [2.0,2.8] & 2.5 & 3.6 & 100.0 & 1.7 [0.9,2.3] & 1.9 & 3.2 & 100.0 \\
\hline
All & 1.7 [1.1,2.1] & 1.8 & 3.6 & 100.0 & 1.8 [1.5,2.6] & 2.1 & 3.9 & 100.0 \\
\hline
\end{tabular}}
\end{subtable}

\vspace{0.8ex}

\begin{subtable}{\linewidth}
\centering
\caption{Guidance (margin relocalization): verbal vs. AR.}
\resizebox{\linewidth}{!}{%
\begin{tabular}{l|cccc|cccc}
\hline
 & \multicolumn{4}{c|}{Verbal guidance} &
   \multicolumn{4}{c}{AR guidance} \\
User &
Med [Q1,Q3] & RMSE & Haus & Cov$_5$ &
Med [Q1,Q3] & RMSE & Haus & Cov$_5$ \\
\hline
P1 & 11.0 [9.4,11.7] & 11.2 & 14.7 & 0.0 & 2.5 [2.1,2.8] & 2.6 & 3.6 & 100.0 \\
P2 & 13.3 [9.5,16.4] & 13.9 & 19.0 & 0.0 & 3.3 [2.7,3.9] & 3.3 & 4.0 & 100.0 \\
P3 & 12.6 [6.5,18.0] & 14.1 & 21.4 & 33.3 & 3.0 [2.7,3.6] & 3.3 & 4.7 & 100.0 \\
P4 & 17.3 [16.0,23.0] & 20.3 & 28.6 & 0.0 & 4.4 [4.1,4.4] & 4.2 & 4.6 & 100.0 \\
P5 & 9.4 [8.8,13.3] & 11.4 & 16.7 & 0.0 & 3.6 [2.9,3.9] & 3.4 & 4.6 & 100.0 \\
P6 & 21.9 [17.1,22.5] & 20.5 & 24.9 & 0.0 & 2.7 [2.6,2.9] & 2.8 & 3.9 & 100.0 \\
\hline
All & 14.2 [9.6,17.8] & 15.7 & 28.6 & 5.6 & 3.2 [2.6,3.9] & 3.3 & 4.7 & 100.0 \\
\hline
\end{tabular}}
\end{subtable}

\end{table}

\subsubsection{Margin relocalization task performance}
We evaluated margin relocalization using $6$ margin targets per user ($6$ users), yielding $36$ localization attempts per condition. Figure~\ref{fig:fig1_violin}(b) shows the localization error distributions for conventional verbal guidance versus AR-guided localization.

With verbal guidance, users often mislocalized the margin (median $14.2$~mm; Q1--Q3: $9.6$--$17.8$~mm), with a maximum error of $28.6$~mm (Table~\ref{tab:big_results_table}). With AR rendering indicating the target, errors were drastically reduced (median $3.2$~mm; Q1--Q3: $2.6$--$3.9$~mm), and all AR-guided attempts fell within $5$~mm (Cov$_5$ = $100\%$). In contrast, only $5.6\%$ of verbal attempts fell within $5$~mm. AR guidance improved localization accuracy for every participant, with a mean reduction of $11.0$~mm in participant-level median error (paired $t$-test on participant-level medians, $p=0.002$). Nearly half of verbal attempts exceeded $15$~mm error, while all AR-guided attempts remained below $5$~mm.

\begin{figure}[t]
\centering    
\includegraphics[width=1\linewidth]{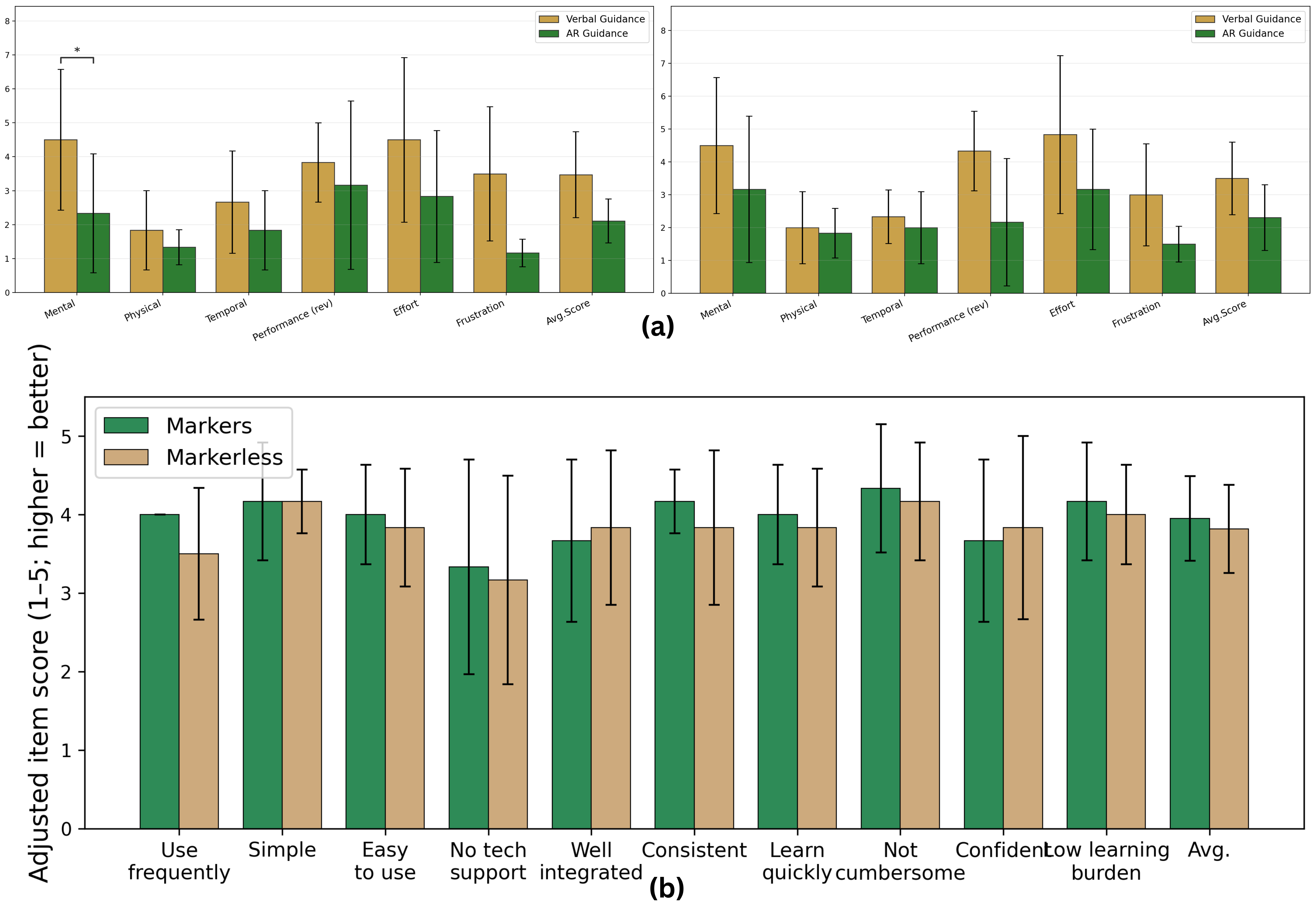}
\caption{(a) NASA-TLX for margin relocalization. (b) System Usability Scale.}
\label{fig:tlx}
\end{figure}

\subsubsection{Workload: NASA-TLX \& System Usability Scale (SUS)}
Figure~\ref{fig:tlx}(a) summarizes NASA-TLX subscales for the margin relocalization task under \textit{verbal} versus \textit{AR-guided} localization, stratified by registration condition (Markers vs.\ Markerless). We used the six standard TLX subscales (Mental, Physical, Temporal, Performance, Effort, Frustration). Performance was reverse-coded as $(10-\text{Performance})$ so that higher values always indicate higher workload / worse perceived performance. Avg.Score is the mean of the six subscales. We used an exact paired Wilcoxon signed-rank test (two-sided) comparing Verbal vs.\ AR within each registration condition ($n=6$). In the \textbf{Markers} condition, AR reduced mental demand (4.50$\pm$2.07 vs.\ 2.33$\pm$1.75; $p=0.0313$). Avg.Score and several subscales trended lower with AR but did not reach $p<0.05$. In the \textbf{Markerless} condition, AR again trended toward lower workload (Avg.Score 3.50$\pm$1.11 vs.\ 2.31$\pm$1.00; $p=0.0625$), but no subscale was significant at $\alpha=0.05$.

Figure~\ref{fig:tlx}(b) shows SUS item scores (1--5) after reverse-scoring negative SUS items so that higher always indicates better usability. We also computed SUS Total (0--100) using standard scoring. Overall usability was similar between registration modes: SUS Total was 73.75$\pm$13.49 (Markers) vs.\ 70.42$\pm$14.00 (Markerless), paired Wilcoxon $p=0.4375$. No item-level differences were significant (all Wilcoxon $p \ge 0.159$).

\begin{figure}[t]
\centering
\includegraphics[width=1\linewidth]{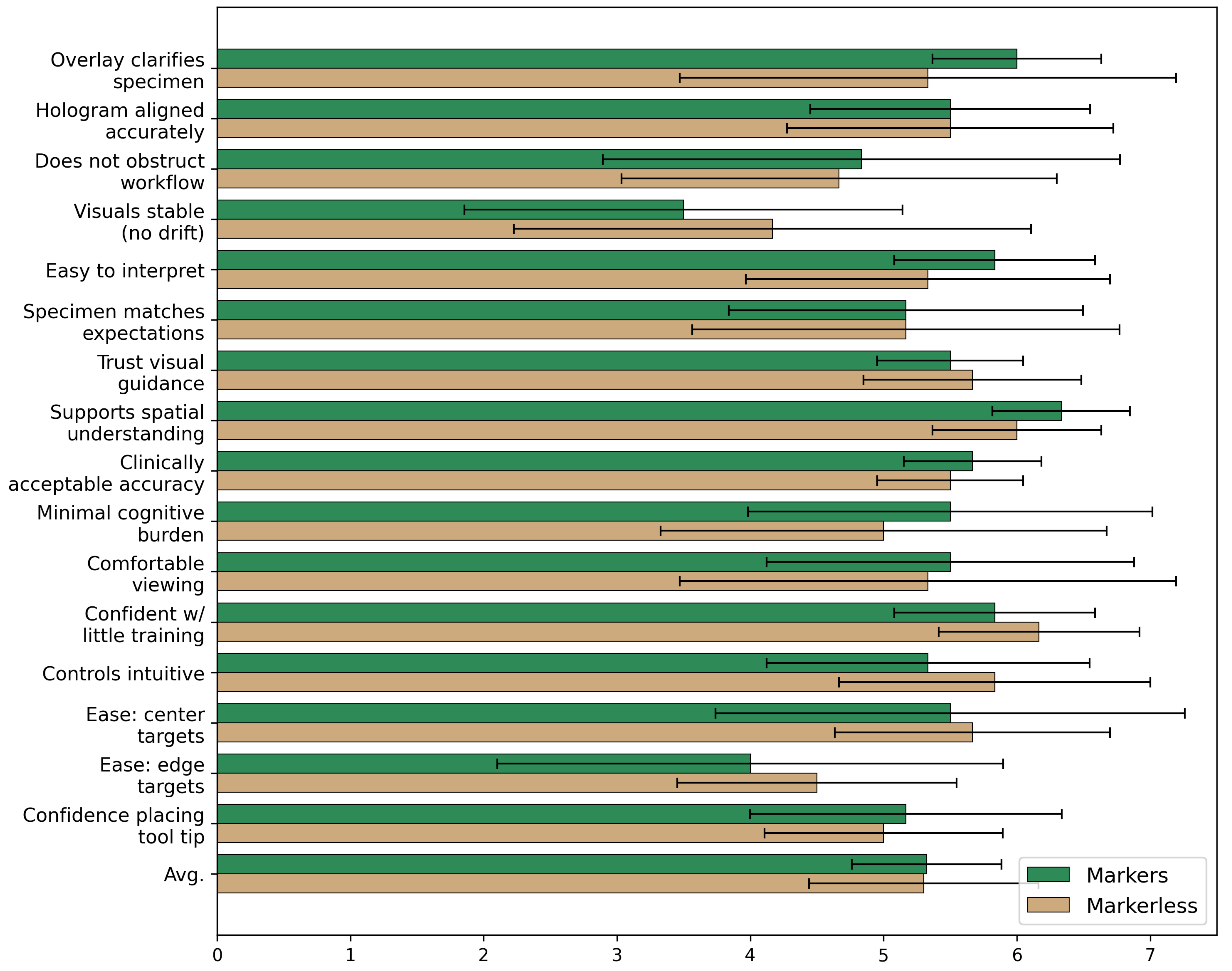}
\caption{AR task-specific ratings}
\label{fig:arq}
\end{figure}

\subsubsection{AR task-specific questionnaire}
Figure~\ref{fig:arq} summarizes the AR task-specific questionnaire, including the overall average across items (AR Avg). We used paired Wilcoxon signed-rank (two-sided) per item and reported Holm-adjusted $p$ values within the questionnaire family. No item or overall average differed significantly between Markers and Markerless (all Wilcoxon $p \ge 0.157$). AR Avg was 5.32$\pm$0.56 (Markers) vs.\ 5.30$\pm$0.86 (Markerless).

\section{Discussion and Conclusion}
We demonstrated a depth sensor-based, markerless AR guidance system for head and neck tumor resection that automatically registers the specimen to the patient and visualizes positive margins in situ. Our study is limited by the form factor of HoloLens, as we noticed reports of discomfort from residents wearing the HMD. However, during the semi-structured interview at end, P1 noted, "people put our loops on and off in the OR and I feel like that's even harder to position correctly. And so I think it'd be very easy for someone to put it [HMD] on your head and take it off when scrubbed-in." With new medical AR device being actively developed, it will not be a critical problem \cite{snke_xr_2026}. Thus, we showed in a controlled phantom study that markerless registration achieved around 2~mm accuracy, and AR guidance improved margin relocation from centimeter-level error to a few millimeters. These results suggest potential clinical value by enabling more precise re-excisions when positive margins are detected, which may reduce unnecessary tissue removal as AR becomes more common in the operating room. Future work includes developing deformation modeling to fit the resected specimen back to the resection bed under deformations and tissue shrinkage, and verifying such development in cadaveric and clinical settings. 

\section{Acknowledgment}
We thank the participants for their active enrollment in this study. This work was supported by the National Institutes of Health (NIH) under grant number 1R01EB037685-01.

%
%
%
 \bibliographystyle{splncs04}
 \bibliography{mybib}

\end{document}